\documentclass[letterpaper, 10 pt, conference]{ieeeconf}  % Comment this line out if you need a4paper

\IEEEoverridecommandlockouts                              % This command is only needed if 
\usepackage{amssymb}  % assumes amsmath package installed
\usepackage{gradient-text}

\title{\LARGE \bf

\textsf{\gradientRGB{STaR}{254,50,254}{15,224,238}}: Scalable Task-Conditioned Retrieval for Long-Horizon Multimodal Robot Memory
\vspace{-0.5em}
\author{Anonymous Authors}
\author{Mingfeng Yuan$^{1}$, Hao Zhang$^{2}$, Mahan Mohammadi$^{1}$, Runhao Li$^{1}$, Jinjun Shan$^{2}$ and Steven L. Waslander$^{1}$ 
}
\thanks{$^1$ University of Toronto Institute for Aerospace Studies and the University of Toronto Robotics Institute, Toronto, Canada
	{\tt\small\{mingfeng.yuan, steven.waslander\}@robotics.utias.utoronto.ca}}
\thanks{$^2$ Department of Earth and Space Science, Lassonde School of Engineering, York University, Toronto, Canada.}}

\def\BibTeX{{\rm B\kern-.05em{\sc i\kern-.025em b}\kern-.08em
    T\kern-.1667em\lower.7ex\hbox{E}\kern-.125emX}}
\usepackage{balance}
\usepackage{amsmath,amsfonts}
\usepackage{array}
\usepackage{textcomp}
\usepackage{stfloats}
\usepackage{url}
\usepackage{verbatim}
\usepackage{graphicx}
\usepackage{cite}
\usepackage[dvipsnames]{xcolor}
\usepackage{caption, subcaption}
\usepackage{float}
\usepackage{diagbox}
\usepackage{caption}%%??captionof??
\usepackage{booktabs}
\usepackage{epstopdf}
\usepackage{bm}
\usepackage{hyperref}
\usepackage{multirow}
\usepackage{cuted}%%\stripsep-3pt
\usepackage{booktabs} % For professional-looking tables
\usepackage{afterpage}  % Ensures figure appears right after 
\usepackage{xcolor}

\usepackage{algorithm}
\usepackage{algpseudocode}
\usepackage{amsmath}
\bstctlcite{IEEEexample:BSTcontrol}

\begin{document}

\makeatletter
\let\@oldmaketitle\@maketitle
\renewcommand{\@maketitle}{\@oldmaketitle
\centering
\vspace{0.5em}
\includegraphics[width=0.95\linewidth, trim={1.1cm 0 0.8cm 0}]{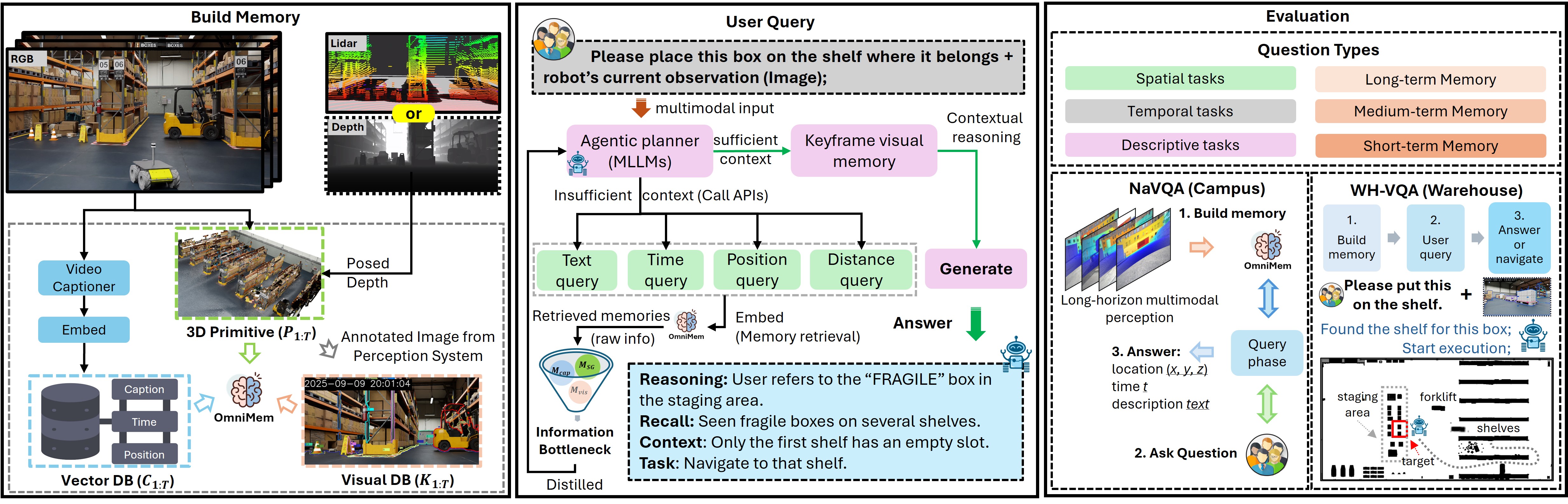}

% The query question is a little bit vague for the reader: Put this box from the staging area where it belongs
% \vspace{-1.5em}
\captionof{figure}{
% \gradientRGB{OpenNav}{254,50,254}{15,224,238} Capabilities - 
\footnotesize \textbf{\emph{STaR} System Overview}.
Our framework consists of three stages. (Left) Memory construction: the robot records RGB and posed depth data to build a multimodal memory composed of three complementary databases (DB) -- video caption, 3D primitive, and visual keyframe -- jointly forming OmniMem.
(Middle) User query and reasoning: given text or multimodal queries, an agentic planner (MLLM) retrieves task-relevant memories through an Information Bottleneck, performs contextual reasoning, and outputs structured answers (location, time, or description).
(Right) Evaluation: We evaluate STaR on both the NaVQA dataset (campus) and the WH-VQA dataset (warehouse), which cover spatial, temporal, and descriptive question types across short-, medium-, and long-term memory settings. The evaluation examines three key capabilities-long horizon cross-modal memory construction, task-conditioned memory retrieval, and contextual reasoning. We also validate the multi-modal query and navigation tasks in a warehouse simulated with Isaac Sim.} % We compare STaR method with existing approaches, sequentially performing memory construction and question answering.
\label{fig:framework}
\vspace{-0.98em}
}
\makeatother

\maketitle\
% \pagestyle{fancy}
% \fancyhf{}
% \renewcommand{\headrulewidth}{0pt}
% \fancyhead[L]{\footnotesize 2025 IEEE/RSJ International Conference on Intelligent Robots and Systems (IROS)\\ October 19-25, 2025. Hangzhou, China.}
% \thispagestyle{fancy}
\thispagestyle{empty}
\pagestyle{empty}
%%%%%%%%%%%%%%%%%%%%%%%%%%%%%%%%%%%%%%%%%%%%%%%%%%%%%%%%%%%%%%%%%%%%%%%%%%
%%%%%%

\begin{abstract}
Mobile robots are often deployed over long durations in diverse open, dynamic scenes, including indoor setting such as warehouses and manufacturing facilities, and outdoor settings such as agricultural and roadway operations. A core challenge is to build a scalable long-horizon memory that supports an agentic workflow for planning, retrieval, and reasoning over open-ended instructions at variable granularity, while producing precise, actionable answers for navigation. We present \textbf{\emph{STaR}}, an agentic reasoning framework that (i) constructs a task-agnostic, multimodal long-term memory that generalizes to \emph{unseen} queries while preserving \emph{fine-grained} environmental semantics (object attributes, spatial relations, and dynamic events), and (ii) introduces a Scalable Task-Conditioned Retrieval algorithm based on the Information Bottleneck principle to extract from long-term memory a compact, non-redundant, information-rich set of candidate memories for contextual reasoning. We evaluate \emph{STaR} on \textbf{NaVQA} (mixed indoor/outdoor campus scenes) and WH-VQA, a customized warehouse benchmark with many visually similar objects built with \textbf{Isaac Sim}, emphasizing contextual reasoning. Across the two datasets, \emph{STaR} consistently outperforms strong baselines, achieving higher success rates and markedly lower spatial error. We further deploy \emph{STaR} on a real Husky wheeled robot in both indoor and outdoor environments, demonstrating robust long-horizon reasoning, scalability, and practical utility. \noindent\textbf{Project Website:} \href{https://trailab.github.io/STaR-website/}{https://trailab.github.io/STaR-website/}

\end{abstract}
\vspace{-0.1cm}
%%%%%%%%%%%%%%%%%%%%%%%%%%%%%%%%%%%%%%%%%%%%%%%%%%%%%%%%%%%%%%%%%%%%%%%%%%%%%%%%
% \input{section/intro1}
% \input{section/intro2}
\section{INTRODUCTION}
Robots are increasingly deployed across indoor environments such as hospitals, manufacturing facilities, and warehouses, as well as outdoor settings, such as driving environments, agricultural lands and mining operations. 
With the recent and rapid progress of multimodal large language models (MLLMs), the abilities of robots to perceive their surroundings and to understand natural language have shifted dramatically into the possible. 
However, a central challenge in embodied intelligence that remains is enabling robots to build long-horizon, extensible spatio-temporal memory in open-world environments so that they can answer free-form human questions and execute multi-step tasks. Such tasks often hinge on variable object granularity, dynamic events, and large numbers of visually similar objects---signals that are hard to capture with conventional metric or semantic maps yet are crucial for real-world operation. The robot must distill a compact, non-redundant, information-rich representation of its environment from hours to days of historical sensor data, and then perform temporal, semantic, and spatial contextual reasoning to resolve fine distinctions, answer complex questions precisely, and reliably execute downstream behaviors such as navigation and manipulation.
% placeholder for comments

To confront open-world variability, a common approach is to construct open-vocabulary 3D scene graphs with large foundation models. Class-agnostic segmentation\cite{gu2024conceptgraphs, deng2024opengraph} (e.g., Segment Anything, Tokenize Anything (TAP)) yields fine-grained segments where each segment is embedded with feature representations from models such as CLIP, and objects are formed by grouping segments using a combination of weighted geometric overlap and semantic similarity. 
However, merging 3D primitives into object-level representations remains insufficient for open-ended tasks with variable granularity. For example, in a warehouse, a request such as “retrieve a box from shelf with a fragile label” may correspond to many visually similar boxes distributed across staging areas, shelves, and unloading zones; moreover, dozens of such boxes may be stacked on a pallet. If retrieval uses “object” as the smallest unit, the top-k results can easily collapse into a single local region (e.g., one pallet), losing broader environmental context that is critical for disambiguation.
%In warehouse settings, the same query (e.g., “fragile box”) may require global aggregation, such as identifying multiple pallets stacked with fragile boxes during inbound logistics, or fine-grained resolution, such as selecting a graspable box from the top of a stacked pallet during order fulfillment. Fixed object-level retrieval therefore either collapses global context or fails to support local manipulability.
While Clio~\cite{Maggio2024Clio} represents an important step toward task-driven scene-graph construction, it relies on a predefined task list and repeatedly re-aggregates all primitives for each specification, which may introduce additional computational overhead as primitive counts grow and task requirements evolve. Moreover, most existing scene-graph frameworks remain largely object-centric and static, leaving rich inter-object spatial relations, dynamic events, and variable information granularity insufficiently addressed.
%While \emph{Clio}~\cite{Maggio2024Clio} advances task-driven scene-graph construction, it assumes a predefined task list and repeatedly re-aggregates \emph{all} primitives per task list. As primitive counts grow and tasks change frequently, this  becomes computationally prohibitive and poorly scalable. Moreover, most current scene-graph frameworks concentrate on largely static, object-level representations, leaving rich inter-object spatial relations, dynamic events, and variable information granularity insufficiently addressed.

%However, such pipelines require hand-tuned similarity thresholds to control object counts and merge rules, a challenging yet critical stage in large-scale scene reconstruction. More fundamentally, as highlighted by \emph{Clio}~\cite{Maggio2024Clio}, semantic selection in a map is not driven solely by appearance similarity but by task needs. For example, in a warehouse, if the task is to move a pallet bearing a stack of boxes, the pallet and boxes should be treated as a single unit during memory retrieval and manipulation; if the task is to distinguish ordinary boxes from those marked “fragile” on the same pallet, the representation must instead resolve individual boxes.
A parallel line of work frames long-term robotic memory as long-video question answering. However, most spatiotemporal memory systems are limited to $\sim$1–2 minutes of history due to prohibitive Transformer inference and storage costs. Retrieval-augmented agents such as \emph{ReMEmbR}~\cite{anwar2025remembr} extend querying over text, space, and time to tens of seconds, but purely rely on video caption streams. While multi-frame VLMs yield richer descriptions of dynamics and layout, the resulting memories become highly redundant in long-term deployments, diluting task-relevant evidence and degrading performance on context-dependent reasoning tasks.
%A parallel line of work casts long-term robotic memory as long-video question answering. Current spatiotemporal memory systems typically handle only $\sim$1–2 minutes of history before Transformer inference and storage become prohibitive. Retrieval-augmented agent systems such as \emph{ReMEmbR}~\cite{anwar2025remembr} extend queries over text, space, and time to tens of minutes, but rely primarily on fixed-interval video caption streams. Although multi-frame VLMs improve descriptions of dynamics, fine details, and layout, the resulting memory is highly redundant in long-term deployments, diluting relevant evidence and reducing downstream success on tasks that require contextual reasoning.

% I removed (first, second, third)...
%Concretely, the robot records task-agnostic 3D primitives (numerous 3D segment fragments), captures dynamic events and spatial relations, and supports variable granularity so details omitted during construction can be recovered at query time.
\textbf{Contributions}: Our first contribution is a framework for building \emph{long-horizon, multimodal spatio-temporal memory} to support robot operation in open-world environments. Because human instructions are open-ended, the memory must remain broadly useful while preserving sufficient detail to support future, unseen tasks. We build a unified memory, \textbf{OmniMem}, comprising: (i) 3D primitives (geometry, class-level semantics); (ii) temporally aligned video captions that provide mid-level dynamic scene descriptions; and (iii) keyframe visual memory for fine-grained cues. This design supports joint temporal, semantic, and spatial reasoning.

Our second contribution is \textbf{STaR} (\textbf{S}calable \textbf{Ta}sk-conditioned \textbf{R}etrieval via Information Bottleneck (IB)) for task-driven 3D scene understanding and memory retrieval. Unlike naïve Retrieval-Augmented Generation (RAG), which concatenates large amounts of redundant and task-irrelevant memory into an LLM prompt, increasing the risk of hallucinations, STaR applies the IB~\cite{tishby2000information} principle to aggregate primitives and select a non-redundant evidence set.
This yields small yet informative memory subsets that preserve answer accuracy while substantially reducing retrieval and inference costs. Crucially, STaR does not require a predefined task list or re-aggregation of the entire scene; it supports rapidly changing tasks and operates only on selected keyframe intervals, yielding scalability with memory length.
%Our second contribution is a long-memory information distillation algorithm, which we call \textbf{STaR} (\textbf{S}calable \textbf{Ta}sk-conditioned \textbf{R}etrieval via Information Bottleneck). Unlike naïve Retrieval-Augmented Generation (RAG) that concatenates large, redundant, task-irrelevant memories into an LLM prompt, inviting hallucination, STaR performs task-driven 3D scene understanding under the Information Bottleneck~\cite{tishby2000information} principle to aggregate primitives and select a minimal, non-redundant, task-relevant evidence set.

Our third contribution is an \textbf{Agentic RAG Workflow} that closes the loop from user queries to actions. An MLLM agent first reasons over the user’s multimodal input (text + images), autonomously plans a search strategy, and issues API calls to retrieve candidate memory snippets from OmniMem. The robot then uses STaR-distilled memory to conduct semantic, spatial, and temporal reasoning, enabling precise answers and reliable execution of navigation.

We evaluate STaR on the navigation video question answering dataset NaVQA and on WH-VQA, a warehouse benchmark built in Isaac Sim that targets long-horizon, variable-granularity memory retrieval, and compare it against two baselines. We further validate its practicality through end-to-end real-robot deployment.
%We evaluate STaR on \emph{NaVQA} and \textbf{WH-VQA}, a new long-horizon warehouse benchmark in \emph{Isaac Sim}. Compared to text-based caption and flat scene-graph baselines, STaR achieves higher success rates and substantially lower errors on open-ended queries of varying granularity, while retrieving information-rich, task-relevant memories from long-term redundant histories. We further validate the approach with end-to-end real-robot deployment.
%We evaluate our method on the navigation video question answering dataset \emph{NaVQA} and on \textbf{WH-VQA}, a new warehouse benchmark in \emph{Isaac Sim} that targets long-horizon, variable-granularity memory-retrieval tasks, validating our approach’s effectiveness in contextual reasoning. Compared with two text-based memory baselines based on video captions and flat scene-graph memory representations, STaR achieves higher success rates and substantially lower errors in accurately answering open-ended questions of varying granularity. Moreover, it retrieves task-relevant, information-rich memories from long-term redundant memory, demonstrating strong advantages in cross-modal contextual reasoning.
\section{Related works}
\textbf{Question Answering (QA)}. 
Embodied Question Answering extends video QA to interactive, egocentric settings with active evidence gathering~\cite{majumdar2024openeqa, wijmans2019embodied, chandrasegaran2024hourvideo}. T*~\cite{ye2025re} answers open-ended questions over hour-long videos via VLM-based keyframe selection but remains video-only, while ReMEmbR~\cite{anwar2025remembr} maps navigation QA to robot-centric outputs yet suffers from redundant caption-only memory under revisits. Our method instead constructs a multimodal long-term memory with task-conditioned aggregation, enabling non-redundant retrieval and adaptive-granularity reasoning for open-ended queries.

\textbf{Long-Term Spatio-Temporal Memory}. 
Scene graphs have become a dominant robotic memory via open-vocabulary perception, spanning flat and hierarchical indoor representations~\cite{gu2024conceptgraphs, werby2024hierarchical} and outdoor extensions~\cite{deng2024opengraph}. While recent work such as Khronos~\cite{schmid2024khronos} incorporates dynamics, most approaches remain object-centric and largely static, with relations defined mainly by proximity and limited modeling of contextual semantics. Our method addresses these limitations through multimodal context integration and temporal reasoning for fine-grained long-horizon queries.

%Open-vocabulary perception has made scene graphs a central memory representation for robots. Representative indoor works include ConceptGraph (flat)~\cite{gu2024conceptgraphs} and HOV-SG (hierarchical)~\cite{werby2024hierarchical}; for outdoor scenes, OpenGraph builds hierarchical graphs~\cite{deng2024opengraph}. Recent efforts move from static scenes to dynamic ones, e.g., Khronos~\cite{schmid2024khronos}, which maintains scene graphs by tracking short-term dynamics and updating long-term changes. Yet semantics often remain object-level, edges are commonly defined by proximity/contact only, and crucial relational details and dynamic events are underrepresented. Our work directly targets these gaps by coupling object-level structure with fine-grained contextual cues and by reasoning over temporal dynamics to answer granular queries.

\textbf{Task-Driven Representations}. 
Another promising direction is task-shaped memory. Clio~\cite{Maggio2024Clio} applies IB to cluster 3D primitives based on a predefined task set, aligning map granularity to task needs; however, modifying the task list requires reclustering all primitives, limiting scalability. ASHiTA~\cite{chang2025ashita} employs hierarchical IB to decompose high-level instructions into sub-tasks and associate them with relevant 3D primitives. Our approach (i) requires no predefined task list, supporting open-ended queries, and (ii) performs task-conditioned aggregation on only a subset of primitives, substantially reducing computation while maintaining accuracy, enabling scalability.
%Another promising direction is to shape memory based on the task. Clio~\cite{Maggio2024Clio} applies the IB to cluster 3D primitives according to a predefined list of everyday tasks, aligning map granularity to task needs; however, updating the task list requires reclustering all primitives, raising concerns about scalability as the number of primitives grows. ASHiTA~\cite{chang2025ashita} introduces hierarchical IB to decompose high-level instructions into sub-tasks and associate them with relevant 3D primitives, but still assumes an explicit task structure. Our approach differs in two key ways: (i) it does not require a predefined task list, enabling open-ended queries; and (ii) it performs task-conditioned aggregation only on a subset of the 3D primitives, substantially reducing computational cost while preserving accuracy, making the approach practical at scale.

\section{Problem Formulation}
Inspired by ReMEmbR and T*, we formulate our problem as a variant of hour-long video question answering for embodied robots. 
During operation, a robot accumulates a monotonically growing multimodal history \(M_{1:T}\) of onboard sensor data (RGB, depth) with associated pose estimates.
At arbitrary times, a user issues an open-ended query $Q$, whose answer may depend on objects, places, or events across varying spatial and temporal granularities. The goal is to predict an answer $A$ by modeling $p(A \mid Q, M)$.
%Given an open-ended query \(Q\), the goal is to predict an answer \(A\) by modeling \(p(A\,|\,Q, M)\). 
%In practice, the robot operates continuously in an \emph{open-world}, collecting a long-horizon multimodal memory spanning tens of minutes to days. 
%At arbitrary times, a user issues a query whose answer may depend on objects, places, or events across varying spatial and temporal granularities. 
In practice, passing the entire memory to a large multimodal model is both computationally prohibitive and prone to hallucinations from task-irrelevant context. Therefore, this work focuses on three core challenges:
\begin{itemize}
    \item\textbf{Memory construction:} How can a robot build a general, extensible memory that supports arbitrary queries?
    %\item \textbf{Memory construction:} How can a robot build a general and extensible memory representation that supports arbitrary queries?
    \item \textbf{Scalable task-conditioned retrieval:} Given a growing memory \(M_{1:T}\), how can robots extract an optimal subset \(R^* \subseteq M_{1:T}\) that is task-relevant, information-rich, and non-redundant, yet remains information-equivalent to the full memory w.r.t.\ the query, while keeping retrieval efficient as \(T\) increases.
    \item \textbf{Contextual reasoning:} How can a robot perform contextual reasoning to accurately answer open-ended queries in complex, dynamic environments? 
    %\textbf{Contextual reasoning:} How can robots be endowed with contextual reasoning capabilities to accurately answer open-ended user queries in complex, dynamic environments?
\end{itemize}

% \textbf{Memory Representation.}
% We construct a multimodal memory consisting of three complementary components, each time- and pose-stamped:
% \begin{itemize}[leftmargin=*]
%     \item \emph{Task-agnostic 3D primitives:} Over-segmented regions projected to 3D with geometry, pose, and coarse semantics (e.g., CLIP/TAP features).  
%     \item \emph{Multi-frame VLM captions:} Mid-granularity descriptions capturing layout, interactions, and dynamics, with redundancy control over time.  
%     \item \emph{Key-frame visuals:} Fine-granularity evidence balancing storage cost and fidelity.
% \end{itemize}
%Given a multimodal, long-horizon memory, the robot must isolate the minimal task-relevant memory set needed to answer open-ended queries. 
In this work, we consider four query types:
For \textbf{spatial tasks}, such as “Which shelf should the FRAGILE box go on?” The robot must (i) discriminate the fine-grained target (fragile label, not other boxes), (ii) infer shelf availability (free slots), and (iii) output a 3D target pose suitable for navigation.
For \textbf{temporal tasks}, such as “When did you last see the forklift?”, the robot may have encountered it multiple times during exploration and must retrieve all relevant timestamps, returning the most recent one as a time-to-now answer (e.g., “8 mins ago”).
For \textbf{descriptive tasks}, common warehouse questions include: “Does the shelf storing blue barrels still have empty slots?”, “How many additional pallets can that shelf hold?”, or “What was the forklift last carrying?” These tasks demand that the robot retrieve information that is both detailed and precise.
For \textbf{multimodal queries}, human tasks are often underspecified, e.g., pointing to an item in the staging area and saying, “Put this on the correct shelf.” The robot must fuse the ambiguous linguistic cue with the current observation to ground “this” correctly and complete the task.

\textbf{Benchmark.} To stress long-horizon memory and open-world reasoning, we present a warehouse video question answering challenge (WH-VQA) in Isaac Sim and evaluate with our proposed \emph{STaR} pipeline that retrieves, reasons, and outputs actions from the above representation.

\begin{figure*}[!t] 
    \vspace{0.2cm}
	\centering
	\includegraphics[width=0.95\linewidth]{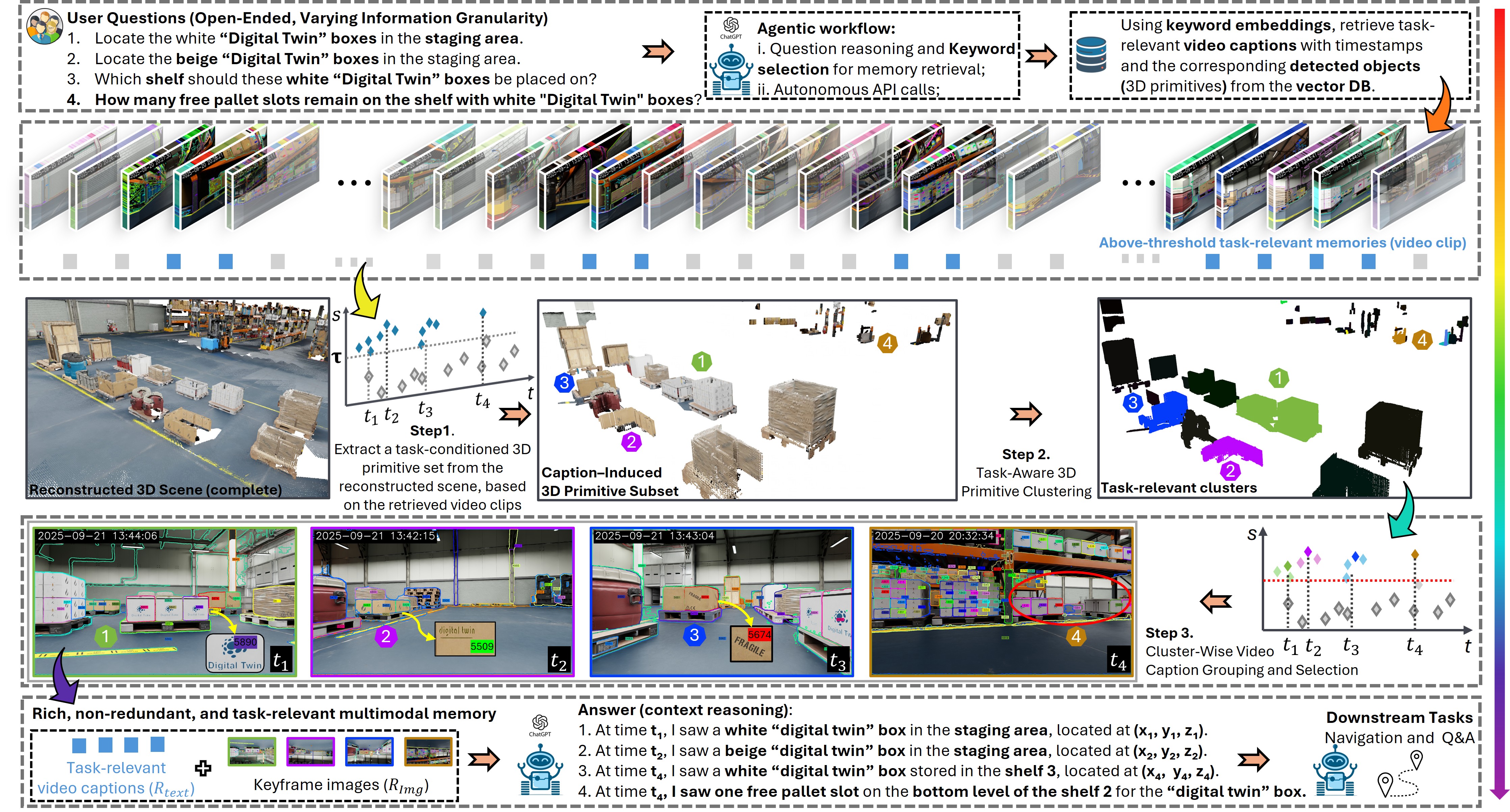}
    \vspace{-0.2cm}
    \caption{\footnotesize \textbf{Task-conditioned retrieval and contextual reasoning}. Given an open-ended query, we embed its cues and query the DB to retrieve \emph{above-threshold} video captions with timestamps and detected objects (Caption–Induced Primitive). These captions induce a working set of primitives \(\mathcal{X}'_Q\), on which we run \emph{IB} to merge \emph{neighboring} primitives into compact, task-relevant clusters. We then group captions by cluster and select one representative caption per cluster to form a non-redundant evidence set. From these memories, the robot optionally loads keyframe images to resolve fine-grained details, performs contextual reasoning, and outputs actionable answers---e.g., locations of white “Digital Twin” boxes in the staging area and shelf indices with remaining pallet slots—supporting navigation and Q\&A.}
    \vspace{-0.6cm}
	\label{fig:AIB}
\end{figure*}

\section{Methodology}
\label{sec:method}
In practice, the optimal task-aware memory subset \(R^*\) is not accessible, so we estimate a subset \(R\) that preserves the information carried by \(R^*\).
We realize this via a task-aware sampler \(D\!:\! M \rightarrow R\) with \(D(M) \subseteq M_{1:T}\), and write
\begin{equation}
p\left(A \mid M_{1: T}, Q\right)=p\left(A \mid R^*, Q\right) \approx p(A \mid R, Q),
\end{equation}
s.t.\ \(R\sim D(M)\).
Our objective is to estimate \(R^*\) so that answers inferred from \(R\) agree with those from the full history \(M\).
Accordingly, we minimize the size of \(R\) while enforcing answer equivalence:
\begin{equation}
R^*=\underset{R}{\operatorname{argmin}}|R|,
\end{equation}
s.t. $\underset{A}{\operatorname{argmax}} p(A \mid R, Q)=\underset{A^{\prime}}{\operatorname{argmax}} p\left(A^{\prime} \mid M, Q\right)$.
% \begin{equation}
% \text { s.t. } \underset{A}{\operatorname{argmax}}\, p(A \mid R, Q)=\underset{A^{\prime}}{\operatorname{argmax}}\, p\left(A^{\prime} \mid M, Q\right).
% \end{equation}
This formulation preserves answer fidelity while enabling compact, task-conditioned retrieval from long-term memory.

Next, we describe how \textbf{\emph{STaR}} constructs the memory \(M\) by aggregating multi-modal observations, and how, at inference time, it samples from \(M\) a compact evidence set \(R \sim \mathcal{D}(M)\).

\subsection{Building OmniMem}
\label{sec:memory-building}
We decompose our multimodal OmniMem as
\(
M_{1:T} = \big(\mathcal{C}_{1:T},\, \mathcal{P}_{1:T},\, \mathcal{K}_{1:T}\big),
\)
where \(\mathcal{C}_{1:T}\) are temporal semantic video captions, \(\mathcal{X}_{1:T}\) are 3D primitives, and \(\mathcal{K}_{1:T}\) are keyframe visual memories, see \textbf{Fig.~\ref{fig:framework}} (left).
Crucially, the queryable memory must be constructed without knowing the user’s future query \(Q\); hence it must remain task-agnostic yet expressive enough to support arbitrary downstream questions.

\noindent \textbf{Video Caption Memory (\(\mathcal{C}_{1:T}\))}.
Following the query-agnostic principle, the robot continuously describes the scene it observes in language.
Specifically, we use NVILA~\cite{liu2025nvila} to caption every \(M\)-second clip, producing a sentence-level description that encodes objects, attributes, events, and spatial relations.
Each caption \(c_t \in \mathcal{C}_{1:T}\) is encoded with a text encoder into an embedding.
We index all embeddings in a \textbf{Milvus} vector database \(\mathcal{V}\),
together with metadata (timestamp, pose, and video caption. %\(\mathbf{z}_t \in \mathbb{R}^d\) \(t\)  \(\mathbf{p}_t\)  \(c_t\)) (e.g., \texttt{mxbai-embed-large-v1}~\cite{Open2024})
The database enables retrieval by text, time, and position.

%The database supports scalable approximate-nearest-neighbor search, enabling retrieval by \textbf{text}, \textbf{time}, and \textbf{position}.

% Each caption \(c_t \in \mathcal{C}_{1:T}\) is embedded with a text encoder (e.g., \texttt{mxbai-embed-large-v1}~\cite{mxbai}) to a vector \(z_t \in \mathbb{R}^d\).
% We maintain \(V\) as a vector index that stores \((z_t,\, \text{timestamp } t,\, \text{pose } P(t),\, \text{aux})\) and supports approximate-nearest-neighbor (ANN) lookups at scale.
% This representation provides a high-recall entry point for open-ended retrieval while preserving temporal and spatial metadata.

\noindent \textbf{Geometric Primitive Memory (\(\mathcal{X}_{1:T}\)).}
To capture persistent structure, We follow the OpenGraph pipeline~\cite{deng2024opengraph} for online 3D scene reconstruction from synchronized RGB, Depth (Lidar/RGB-D), and pose streams.
We adopt an open-vocabulary perception stack combining RAM~\cite{zhang2024recognize}, Grounding-DINO~\cite{liu2024grounding}, and TAP~\cite{pan2024tokenize} models, yielding per-frame masks and captions.
LiDAR scans are filtered by 4DMOS~\cite{mersch2022ral} to remove dynamic objects.
Given camera intrinsics \(K\) and extrinsics \(T\). We back-project point clouds into the 2D mask space to recover object point sets.
DBSCAN is applied to further denoise the
projected subset.
With robot pose, 
%each object cloud is fused into a global frame to form/update a 3D primitive.
we then initially form objects by grouping segments using a weighted combination of geometric overlap and semantic similarity.
Each primitive thus stores: (i) geometry (point cloud), (ii) a
caption, (iii) a semantic feature vector, and (iv) a
list of detection timestamps that allows the system to later link and synchronize
the video captions \(C_{1:T}\) with the keyframe memory \(K_{1:T}\).
% Each primitive thus stores: (i) geometry (3D point cloud), (ii) a caption, and (iii) a semantic feature vector, enabling later cross-modal joins with \(\mathcal{C}_{1:T}\) and \(\mathcal{K}_{1:T}\).
%\(\mathcal{T}_{\text{det}}\)

\noindent \textbf{Visual Keyframe Memory (\(\mathcal{K}_{1:T}\)).}
Fine-grained queries often require details that captions or primitives may miss (e.g., “is there an empty parking slot?”).
To retain such visual evidence without excessive storage, we log \emph{annotated keyframes} at a controlled rate (e.g., \(1\,\text{Hz}\)), guided by novelty checks in the incremental mapping pipeline.
Each keyframe \(k\in\mathcal{K}_{1:T}\) stores the raw image with overlaid object contours, each labeled by its global primitive index, together with a timestamp.
This yields an image-retrievable trail with explicit links to primitives, offering fine visual detail when needed while keeping storage growth manageable.
%\noindent \textbf{Visual Keyframe Memory (\(\mathcal{K}_{1:T}\)).}
% Fine-grained queries often require details that captions or primitives alone may omit (e.g., “is there an empty parking slot?”).
% To preserve such visual evidence without prohibitive storage, we log \emph{annotated keyframes} at a controlled rate (e.g., \(1\,\text{Hz}\)) with novelty checks from the incremental mapping pipeline.
% A keyframe \(k\in\mathcal{K}_{1:T}\) stores the raw image overlaid with all object contours, each labeled by its global primitive index, along with a timestamp.
% This process yields an image-retrievable trail with explicit links to primitives, providing arbitrarily fine visual detail when needed while keeping storage growth manageable.
\subsection{Scalable Task-Conditioned Retrieval}
\label{sec:backend-qcib}
Given a query \(Q\) and the long-horizon memory \(M_{1:T}\), we form a compact, non-redundant evidence set \(R\) in three stages, as shown in Fig.~\ref{fig:AIB}: (i) video caption-induced primitive subset extraction, (ii) task-driven clustering via IB on the induced 3D primitive subset, and (iii) cluster-wise caption grouping and representative selection.

\noindent\textbf{Video-Caption–Induced 3D Primitive Subset.}
Given a query \(Q\), we first retrieve a high-recall pool of captions \(\mathcal{C}_Q \subset \mathcal{C}_{1:T}\) by querying the vector database \(\mathcal{V}\) with \(Q\)-related keywords and retaining all captions whose \emph{normalized} similarity exceeds a preset threshold \(\tau\). Each retained caption is associated with a list of global
indices of 3D primitives detected in that video clip; the union of these lists
forms a working set \(\mathcal{X}'_Q \subset \mathcal{X}_{1:T}\), where
\(\mathcal{X}_{1:T}\) denotes the full collection of primitives over time. Primitives in \(\mathcal{X}'_Q\) carry both semantic information and geometric attributes, which together provide a strong inductive bias for clustering. To explicitly encode this bias, we adopt the \emph{IB}~\cite{Maggio2024Clio},
which forms task-relevant clusters by iteratively merging \emph{neighboring}
primitives, thereby allowing the cluster granularity to be adaptively controlled according to the task. %We prefer to merge spatially adjacent segments while avoiding the merging of distant ones.

\noindent\textbf{Task-Driven Clustering via IB.} To obtain a compact, task-relevant evidence set, we run IB on the primitive subset
\(\mathcal{X}'_Q\) (rather than the full \(\mathcal{X}_{1:T}\))
to find a more compact information \(\tilde{\mathcal{X}}'\), representing the task-relevant
concepts that compresses \(\mathcal{X}'_Q\) while remaining maximally
informative about the task \(Y\).
The task-relevant clusters \(\tilde{\mathcal{X}}'\) are defined through the probability distribution \(p(\tilde{x}'\mid x')\), which specifies
the likelihood that a task-agnostic primitive \(x'\) belongs to a cluster
\(\tilde{x}'\). 
%This formulation allows AIB to identify semantically and geometrically coherent groups that are most informative for the given task.
In practice, this strategy preserves the rich spatial and semantic information
about scene objects conveyed by the video captions while greatly reducing the
computational complexity of the memory retrieval process.
We initialize the task-relevant clusters \(\tilde{\mathcal{X}}'\) to the task-agnostic primitives \(\mathcal{X}'_Q\). Starting from singleton clusters, we iteratively merge adjacent clusters \(\tilde{x}'_i, \tilde{x}'_j\) that minimize the task-driven dissimilarity.
% To obtain a compact, task-relevant evidence set, we run the bottom-up
% AIB on the video caption-induced primitive subset
% \(\mathcal{X}'_Q\) (rather than the full \(\mathcal{X}_{1:T}\))
% to find a more compact signal \(\tilde{\mathcal{X}}'\), representing the task-relevant
% concepts that compresses \(\mathcal{X}'_Q\) while remaining maximally
% informative about the task variables \(Y\).
% This strategy preserves the rich spatial and semantic information about scene objects conveyed by the video captions, while greatly reducing the computational complexity of the memory retrieval process.
% Starting from singleton clusters, at each iteration we merge adjacent clusters
% \(\tilde{x}_i,\tilde{x}_j\) that minimize the task-driven dissimilarity.
% To obtain a compact, task-relevant evidence set, we run the bottom-up
% AIB on the video caption-induced primitive subset
% \(\overline{\mathcal{X}}_Q\) (rather than the full \(\mathcal{X}_{1:T}\)),
% which substantially improves computational efficiency.
% Starting from singleton clusters, at each iteration we merge adjacent clusters
% \(\tilde{x}_i,\tilde{x}_j\) that minimize the task-driven dissimilarity
\begin{equation}
d_{i j}=\left(p\left(\tilde{x}'_i\right)+p\left(\tilde{x}'_j\right)\right) D_{\mathrm{JS}}\left(p\left(y \mid \tilde{x}'_i\right), p\left(y \mid \tilde{x}'_j\right)\right),
\end{equation}
where \(D_{\mathrm{JS}}\) is the Jensen-Shannon divergence and \(p(\tilde{x}')\) is the cluster prior, defined as \(p(\tilde{x}') = \frac{1}{N}\),  
where \(N\) denotes the number of primitives in \(\mathcal{X}'_Q\). Merging proceeds greedily while monitoring the fractional loss
of task information \(\delta{(k)}\). To regulate compression, we compute the fractional information-loss after merge \(k^{th}\) cluster,
\begin{equation}
\delta(k)=\frac{I\left(\tilde{\mathcal{X}}'_{(k)} ; Y\right)-I\left(\tilde{\mathcal{X}}'_{(k-1)} ; Y\right)}{I(\mathcal{X}'_Q ; Y)},
\end{equation}
and stop when \(\delta(k)>\bar{\delta}\).
\begin{align}
I(\mathcal{X}'_Q ; Y) = \sum_{x'} p(x')\sum_{y} p(y\!\mid\!x')\,\log\frac{p(y\!\mid\!x')}{p(y)}. \label{eq:mi-plugin}
\end{align}
IB requires the conditional \(p(y\mid x')\) that quantifies the task-relevance of each primitive \(x'\in\mathcal{X}'_Q\). The cosine similarity function \(\varphi(\cdot,\cdot)\) measures how well a primitive (\(f_{x’}\)) aligns with a task cue (\(f_{t}\)), serving as a semantic relevance score. %We embed both primitives (\(f_{x’}\)) and task cues (\(f_{t_j}\)) into a shared representation space.
% \[
% \theta_j(x)=
% \begin{cases}
% \alpha, & j=0 \;\;(\text{null task})\\
% \varphi\!\big(f_x,f_{t_j}\big), & j\in\{1,\dots,m\}
% \end{cases}
% \]
\begin{equation}
\theta_j(x')= \begin{cases}\alpha, & j=0 \text { (null task) } \\ \varphi\left(f_{x'}, f_{t}\right), & j \in\{1, \ldots, m\}\end{cases}
\end{equation}

\noindent \(\alpha\) is a floor that assigns clearly irrelevant primitives to the null task; We set
% \begin{equation}
% p(q \mid x)= \begin{cases}{[1,0]^{\top}} & \text { if } \varphi\left(f_x, f_{t_j}\right)<\alpha \\ \eta \gamma_k \theta(x) & \text { otherwise }\end{cases}
% \end{equation}
\begin{equation}
p\left(y\mid x_i'\right)=\begin{cases}{\left[
1,0, \dots, 0\right]^{\top},} & \text {if } \varphi\left(f_{x'}, f_{t}\right)<\alpha\\
\eta \sum_{l=1}^k \gamma_k\left(\theta\left(x_i'\right)\right), & \text {otherwise }\end{cases}
\end{equation}
We apply the operator \(\gamma_k(\cdot)\), which retains the top-k task scores while zeroing out the rest, ensuring that only the most relevant primitives contribute to clustering. \(\eta\) is normalization constant. This construction highlights the most relevant task cues while suppressing noise, yielding stable \(p(y \mid x')\) for IB merges and producing clusters \(\tilde{\mathcal{X}}'\) compact to \(Q\).

\noindent\textbf{Cluster-Wise Grouping and Selection.}
To extract informative memories from a redundant pool of video captions,
we leverage the task-driven clusters
\(\tilde{\mathcal{X}}' = \{\tilde{x}'_1, \ldots, \tilde{x}'_N\}\)
obtained previously.
Our intuition is that two captions are redundant if they refer to the same cluster. Each video caption \(c(T_i)\) corresponds to the detailed scene description within the \(i^{th}\) \textit{M}-second segment and is associated with a set of
primitive indices \(I(c(T_i))\). To construct an informative and compact memory for contextual reasoning,
we \textbf{group captions by clusters} \(\tilde{\mathcal{X}}'\).
Specifically, each caption \(c(T_i) \in \mathcal{C}_Q\) is assigned to the
caption group \(\mathcal{G}(\tilde{x}'_k)\) corresponding to cluster
\(\tilde{x}'_k\) (\(k \in \{1, \ldots, N\}\)) if their index sets overlap:
\begin{equation}
\mathcal{G}(\tilde{x}'_k) \leftarrow c(T_i)
\;\;\text{iff}\;\;
I(c(T_i)) \cap \tilde{x}'_k \neq \varnothing.
\end{equation}
% A single \emph{representative} caption \(\hat{c}\) is selected from each group that is most semantically relevant to the query \(Q\):
% Across all clusters \(\tilde{x}'_j \in \tilde{\mathcal{X}}'\), we first select
% a representative caption \(\hat{c}_k\) for each cluster, and then rank all
% representatives globally by their semantic relevance
% \(\phi(\hat{c}_k, Q)\) to the query \(Q\),
% where \(\phi(\cdot, \cdot)\) measures caption--query similarity.
% % \begin{equation}
% % \hat{c}_k=\underset{c \in \mathcal{G}\left(\tilde{x}_k^{\prime}\right)}{\arg \max } \phi(c, Q), k ={1, 2, \dots, N} .
% % \end{equation}
% \begin{equation}
% \hat{c}_k =
% \underset{c \in \mathcal{G}(\tilde{x}'_k)}{\arg\max}\;
% \phi(c, Q),
% \quad
% k \in \{1, 2, \ldots, N\}.
% \end{equation}
% \begin{equation}
% R_{text} = \left\{
% \hat{c}_k \mid
% \mathrm{rank}\big(\phi(\hat{c}_k, Q)\big) \le K
% \right\}
% \end{equation}
Across all clusters \(\tilde{x}'_j \in \tilde{\mathcal{X}}'\), we first select
a representative caption \(\hat{c}_k\) for each cluster and then rank all
representatives globally by their semantic relevance
\(\phi(\hat{c}_k, Q)\) to the query \(Q\),
where \(\phi(\cdot, \cdot)\) measures caption-query similarity:
\begin{equation}
\hat{c}_k =
\underset{c \in \mathcal{G}(\tilde{x}'_k)}{\arg\max}\;
\phi(c, Q),
\quad
k \in \{1, 2, \ldots, N\}.
\end{equation}
Finally, we rank all representative captions based on their scores
and retain the top-\(K\) to form the textual memories:
\begin{equation}
R_{\text{text}} =
\left\{
\hat{c}_k \mid
\mathrm{rank}\!\left(\phi(\hat{c}_k, Q)\right) \le K
\right\}.
\end{equation}
% we therefore \textbf{group captions by clusters \(\tilde{\mathcal{X}}'\)}.
% Specifically, a caption \(c(T_i)\in\mathcal{C}_Q\) is assigned to the caption group \(\mathcal{G}(\tilde{x}'_k)\) of cluster
% \(\tilde{x}'_k\) (\(k \in \{1, \ldots, N\}\)) if their index sets overlap:
% \begin{equation}
% c\left(T_i\right) \in \mathcal{G}\left(\tilde{x}_k^{\prime}\right) \text { iff } I\left(c\left(T_i\right)\right) \cap \tilde{x}_k^{\prime} \neq \varnothing .
% \end{equation}

% \[
% \hat{c}
% =
% \operatornamewithlimits{\arg\max}_{c \in \mathcal{C}(\tilde{x}'_j)}
% \phi(c, Q),
% \]
% where \(\phi(\cdot, \cdot)\) denotes the semantic similarity between a caption and the query.
\noindent This process ensures that the final retrieval set \(R\) is both
\textbf{diverse} (one representative per cluster) and
\textbf{task-relevant} (ranked by semantic importance).

When the query requires finer-grained evidence beyond captions, the LLM acts as a selector that samples a \textbf{set of keyframe timestamps}
\(T^* = \{t_1^*, t_2^*, \ldots, t_m^*\}\) from the video based on the temporal information embedded in the textual memories
\(R_{\text{text}}\).
The corresponding keyframe images are retrieved as \(R_{\text{img}}(T^*)\in \mathcal{K}_{1:T}\),
and the final multimodal memory \(R\) is formed by fusing the text and image modalities:
\begin{equation}
R=R_{\text {text }} \oplus R_{\text {img }}\left(T^*\right), \text { where } T^*=f_{\text {LLM}}\left(R_{\text{text }}\right)
\end{equation}
A concise illustration of this pipeline is shown in Fig.~\ref{fig:AIB}.

\section{Experimental Setup}
%\subsection{Dataset}
%Motivated by the real-world challenge that user queries are open-ended and that the granularity of key evidential cues varies, 
%In this section, we present the experimental evaluation of \emph{STaR} on the NaVQA~\cite{anwar2025remembr} and WH-VQA benchmarks introduced above. We compare \textbf{STaR} against two state-of-the-art baselines: the ReMEmbR framework, which uses video-caption-based memory retrieval, and an OpenGraph-based open-vocabulary perception system that produces object-level captions.
We evaluate the proposed \emph{STaR} algorithm on the NaVQA~\cite{anwar2025remembr} and WH-VQA datasets and compare its performance with two state-of-the-art baselines: the ReMEmbR framework, which relies on video-caption-based memory, and an OpenGraph-based open-vocabulary perception system that generates object-level captions.
%We evaluate \emph{STaR} using two complementary benchmarks: NaVQA~\cite{anwar2025remembr} and the WH-VQA dataset. The former focuses on long-horizon navigation question answering, whereas the latter represents a realistic industrial scenario that replicates the redundant memories a robot may accumulate during long-term deployment in a warehouse, as well as its ability to retrieve memories at varying levels of granularity. Together they enable a comprehensive assessment of long-term memory retrieval and contextual reasoning in environments populated with numerous task-relevant, visually similar objects.

\subsubsection{\textbf{NaVQA dataset}} NaVQA is built on the CODa robot navigation dataset~\cite{zhang2024toward} and targets long-horizon video QA for navigation. The data provide time-synchronized RGB, LiDAR, and localization signals collected under varied illumination and times of day across seven long sequences in indoor and outdoor campus environments. Data sequences are partitioned by duration into short ($<2$ min, medium (2–7 min), and long (7–35.9 min) memories. Each split includes three task types, spatial, temporal, and textual with the textual set comprising binary (yes/no) and descriptive questions, totaling 210 queries. Notably, 22 queries fall outside the memory window and are excluded to ensure fairness.
%\color{blue} We partition sequences by duration into short ($\verb|<|$2 min), medium (2--7 min), and long (7--35.9 min) memories. Each split contains three task families, spatial (35\%), temporal (15\%), and textual questions; the textual set includes binary yes/no (33\%) and descriptive (17\%) questions, for a total of 200 questions.
%\textbf{NaVQA dataset}. The NaVQA is built on the CODa robot navigation dataset~\cite{zhang2024toward} and targets long-horizon navigation video QA. The navigation data provide time-synchronized RGB, LiDAR, and localization signals collected under varied illumination and times of day, covering seven long-duration navigation sequences across indoor and outdoor campus environments. We partition sequences by temporal span into short ($\verb|<|$2 min), medium (2–7 min), and long (7–30 min) memories. Each split comprises three task families: spatial (35\%), temporal (15\%), and textual questions, where the textual questions include binary yes/no (33\%) and descriptive (17\%) questions, for a total of 200 questions.

\subsubsection{\textbf{WH-VQA dataset}} We additionally create WH-VQA, a customized warehouse benchmark built in Isaac Sim, and expand it with 100 evaluation tasks. The dataset captures a 22-minute navigation trajectory and is more challenging than NaVQA due to the presence of many visually similar objects that require task-level disambiguation. \color{black}Its task composition mirrors NaVQA---spatial (40\%), binary yes/no (20\%), and descriptive (20\%)---while adding multimodal queries (20\%) to emulate human-robot interaction with underspecified language requiring grounding against the current visual scene. 
\subsubsection{\textbf{Evaluation Metrics}} We design task-specific metrics for each question type.
For \textbf{spatial questions}, which require \((x,y,z)\) outputs, we compute the L2 error between predicted and ground-truth positions and deem answers correct if the error is below 15~m on NaVQA (following ReMEmbR) and 5~m on WH-VQA, reflecting the finer spatial structure of warehouse shelves and pallets.  
For \textbf{temporal questions}, which return answers such as “15 minutes ago,” we use L1 temporal error and mark predictions within 2 minutes as correct.  
For \textbf{textual questions}, yes/no accuracy is used, while descriptive answers are automatically judged by an LLM using both the model output and ground-truth annotation.  
Recall@K ($R^k$) is defined as the fraction of queries for which, after multiple tool-call rounds, the retrieved memory set of size $k$ contains at least one entry whose timestamp falls within a tolerance window (e.g., $\pm 5,\mathrm{s}$) of the ground-truth timestamp.
%We also introduce a \textbf{Recall} metric evaluating whether the algorithm retrieves task-relevant memories: retrieval is correct if the timestamp differs by less than 5~s from the ground-truth context.  
Finally, we report \textbf{overall correctness} by averaging success rates across all question categories.
\color{black}
\section{Results}
%\subsection{Evaluation on the dataset}
\begin{figure*}[!t] 
    \vspace{0.2cm}
	\centering
	\includegraphics[width=0.9\linewidth]{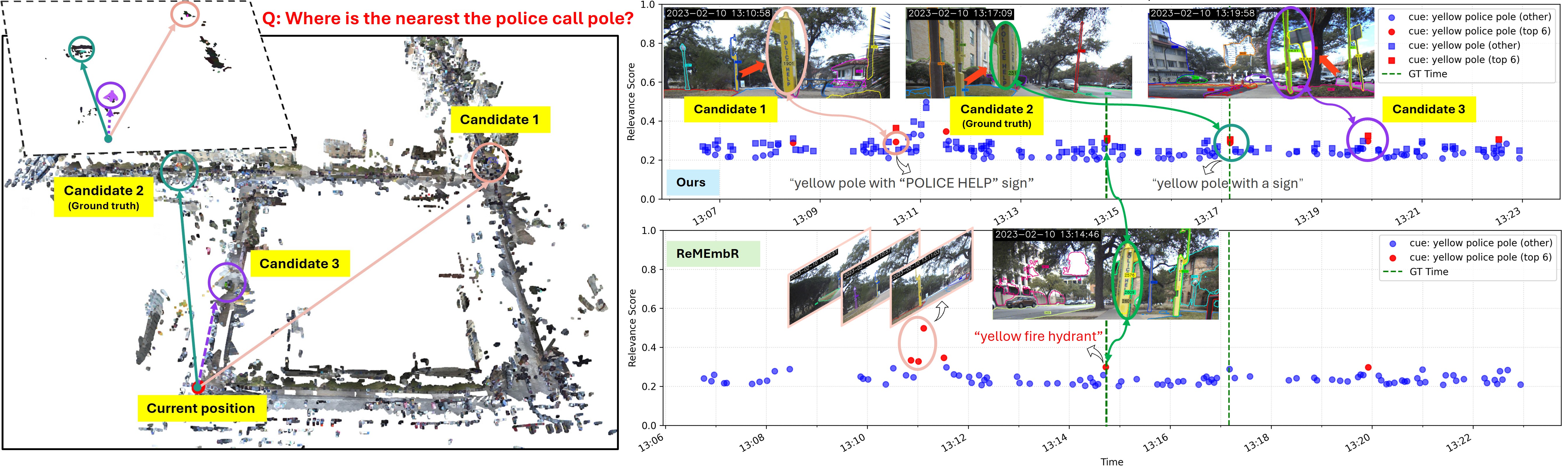}
    \vspace{-0.3cm}
	\caption{\footnotesize \textbf{Qualitative example.} Left: 3D reconstruction for a 16-min memory horizon with the robot’s current pose (red) and three candidate answers. Task-relevant 3D primitives selected by our IB-based clustering are highlighted; task-irrelevant 3D primitives are masked (not shown). Right: temporal retrieval. STaR selects diverse, non-redundant captions and correctly grounds the “yellow pole with POLICE HELP sign,” choosing Candidate 2 as the nearest police pole. In contrast, ReMEmbR (bottom) repeatedly retrieves redundant captions at timestep 13:11 and fails to identify the correct nearest target.}
	\label{fig:police}
    \vspace{-0.5cm}
\end{figure*}
\begin{figure}[tbp] 
	\centering
\includegraphics[width=1\linewidth]{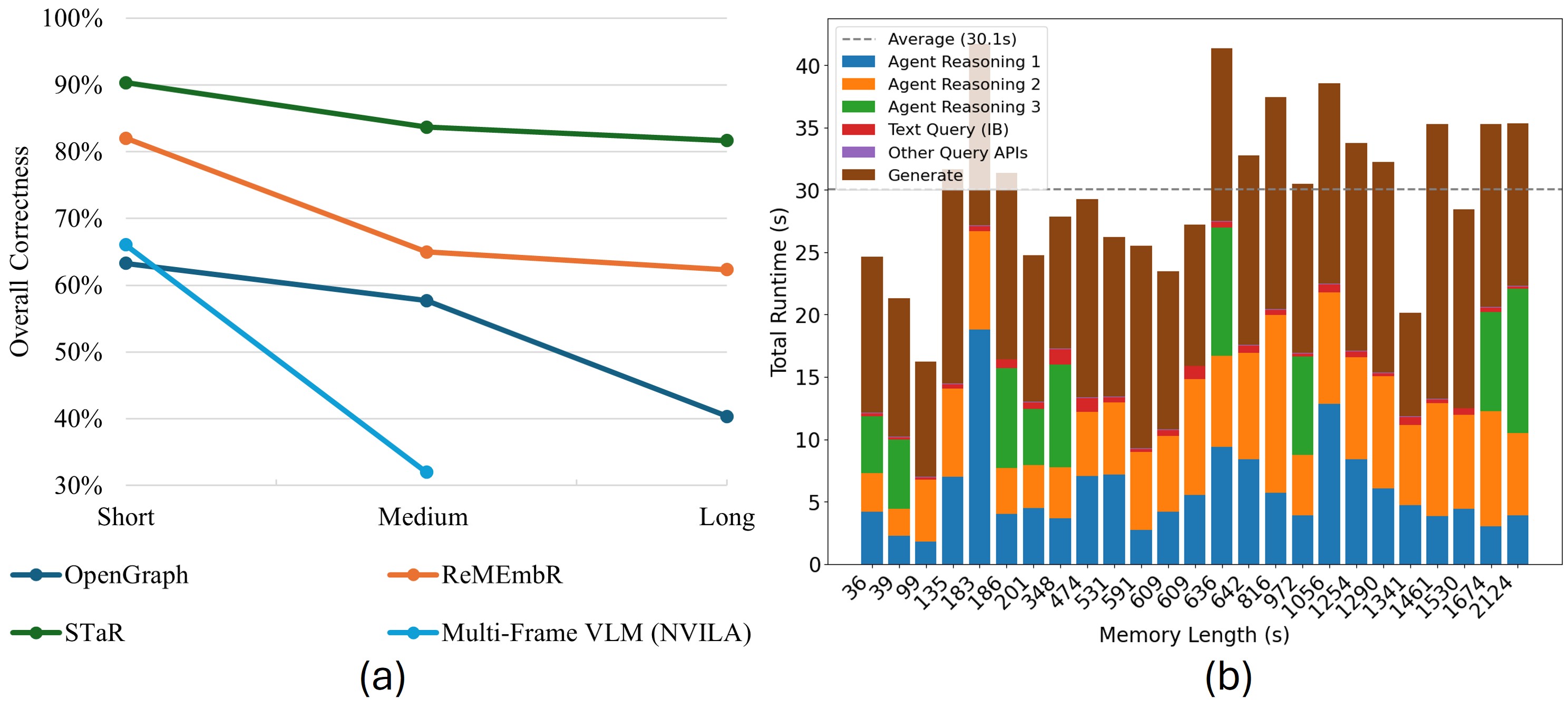}
	\caption{\footnotesize Scalability of STaR on the NaVQA dataset with increasing memory length (from 36 seconds to 35.9 minutes): (a) Overall success rate; (b) Runtime breakdown of STaR.}
	\label{fig:overall}
    \vspace{-0.3cm}
\end{figure}

\begin{table}[t!]
  \centering
  \setlength{\tabcolsep}{2.6pt}
  \renewcommand{\arraystretch}{1.0}
  \begin{tabular}{@{}llccccccccc@{}}
    \toprule
    \multirow{2}{*}{\textbf{Task}} & \multirow{2}{*}{\textbf{Method}} &
    \multicolumn{3}{c}{\textbf{Short}} &
    \multicolumn{3}{c}{\textbf{Medium}} &
    \multicolumn{3}{c}{\textbf{Long}} \\
    \cmidrule(lr){3-5}\cmidrule(lr){6-8}\cmidrule(lr){9-11}
    & &
    \textbf{SR}$\uparrow$ & \textbf{Er.}$\downarrow$ & $\mathrm{\textbf{R}}^k$$\uparrow$ &
    \textbf{SR}$\uparrow$ & \textbf{Er.}$\downarrow$ & $\mathrm{\textbf{R}}^k$$\uparrow$ &
    \textbf{SR}$\uparrow$ & \textbf{Er.}$\downarrow$ & $\mathrm{\textbf{R}}^k$$\uparrow$ \\
    \midrule
    \multirow{3}{*}{\textbf{Spat.}}
      & OpenGraph & 0.68 & 21.1 & 0.74 & 0.63 & 20.5 & 0.73 & 0.26 & 87.4 & 0.72 \\
      & ReMEmbR   & 0.84 & 9.6  & 0.84 & 0.63 & 20.4 & 0.79 & 0.49 & 53.9 & 0.80 \\
      & \textbf{STaR} &
        \textbf{0.89} & \textbf{4.2} & \textbf{0.89} &
        \textbf{0.84} & \textbf{10.8} & \textbf{0.84} &
        \textbf{0.77} & \textbf{16.9} & \textbf{0.84} \\
    \midrule
    \multirow{3}{*}{\textbf{Temp.}}
      & OpenGraph & 0.89 & 4.5 & 0.50 & 0.67 & 5.3 & 0.87 & 0.59 & 7.1 & 0.77 \\
      & ReMEmbR   & \textbf{1.00} & \textbf{0.2} & \textbf{1.00} & 0.67 & 1.7 & 0.67 & \textbf{0.88} & 3.4 & \textbf{0.88} \\
      & \textbf{STaR} &
        \textbf{1.00} & 0.4 & \textbf{1.00} &
        \textbf{0.83} & \textbf{0.9} & \textbf{0.87} &
        \textbf{0.88} & \textbf{2.7} & \textbf{0.88} \\
    \midrule
    \multirow{3}{*}{\textbf{Text}}
      & OpenGraph & 0.33 & -- & 0.61 & 0.43 & -- & 0.67 & 0.36 & -- & 0.69 \\
      & ReMEmbR   & 0.62 & -- & 0.76 & 0.65 & -- & 0.77 & 0.50 & -- & 0.76 \\
      & \textbf{STaR} &
        \textbf{0.82} & -- & \textbf{0.89} &
        \textbf{0.84} & -- & \textbf{0.84} &
        \textbf{0.80} & -- & \textbf{0.80} \\
    \bottomrule
  \end{tabular}

  \caption{\footnotesize
  Comparison on the NaVQA dataset (campus, mixed indoor–outdoor).
  SR: success rate; Er.: error (L2 distance for Spatial (Spat.) tasks and L1 distance for Temporal (Temp.) tasks);
  $R^k$: recall@k -- the top-K retrieved memory entries include the ground-truth timestamp.
  For the Text category, Er. is not applicable and marked “--”.}
  \label{tab:performance1}
  \vspace{-0.3cm}
\end{table}

\subsubsection{\textbf{NaVQA results}} 
%Fig.~\ref{fig:overall}-(a) shows overall correctness across short, medium, and long-horizon memories. As memory grows, STaR maintains a consistent lead over all baselines, particularly in the long-horizon memory. For the multi-frame VLM baseline, which uniformly sample frames from memory at fixed intervals and feeds them to a Transformer-based VLM, performance degrades sharply as the memory length increases. Next, we break down performance by task in Table~\ref{tab:performance1}.
Fig.~\ref{fig:overall} (a) shows overall correctness across short-, medium-, and long-horizon memories. As memory grows, \emph{STaR} consistently outperforms all baselines, particularly in long-horizon settings, while the multi-frame VLM baseline degrades sharply. Runtime scalability is evaluated on 25 NaVQA tasks with memory lengths ranging from 36s to 35.9 minutes, yielding a stable average query time of 30.1~s. As shown in Fig.~\ref{fig:overall} (b) and the agentic RAG workflow in Fig.~\ref{fig:framework}, the agentic planner performs up to three reasoning rounds (5.9~s, 6.8~s, and 2.7~s on average), with later rounds invoked only when earlier retrieval results are insufficient. Answer generation dominates runtime (14.2~s), while Milvus-based memory retrieval remains efficient (text query API $\sim$0.5~s due to information distillation, other query APIs $\leq$0.1~s). Both the agentic planner and the answer generation module use \textit{ChatGPT-4.1-mini}, and our STaR framework operates in a zero-shot setting. Next, we break down query performance by task in Table~\ref{tab:performance1}.

\textbf{Spatial tasks}. We achieve the highest success rates across all horizons with approximately two to four times lower spatial error than the baselines. In contrast, ReMEmbR and OpenGraph show rapidly increasing errors that exceed 50 m on long sequences. 
These gains stem from two main factors: (1) our STaR retrieves compact, task-relevant evidence sets that suppress redundant memories, and (2) our answers are grounded in the 3D bounding-box centers of target objects rather than the robot’s pose at the time of observation, as in ReMEmbR, yielding more accurate and directly actionable navigation goals.
\textbf{Temporal tasks}. All methods perform well, as NaVQA temporal questions involve coarse semantics and simple reasoning. Nonetheless, STaR consistently achieves the highest success rate and smallest temporal errors over longer horizons.
\textbf{Textual tasks}. For descriptive and binary questions, our model achieves the strongest performance, with clear gains over baselines. Unlike ReMEmbR (video captions only) or OpenGraph (object-level captions only), our system reasons over variable-granularity semantics by jointly leveraging captions, 3D primitives, and keyframe imagery. This enables, for example, correctly determining nearby parking availability by retrieving candidate parking-area memories and re-examining keyframes at relevant timestamps, leading to improved performance on long-horizon textual queries requiring fine-grained visual understanding. Across all tasks, baselines show high recall but low success rates due to insufficient task detail in text-only representations, leading to incorrect target selection.
%For descriptive and binary questions, our model achieves the strongest overall performance, improving success rates by over 10 percentage points compared to ReMEmbR and more than doubling those of OpenGraph on long sequences (e.g., 1.00 vs 0.89 and 0.36). Unlike ReMEmbR, which relies purely on video captions, or OpenGraph, which provides only object-level captions, our system can reason over variable-granularity semantics by combining captions, 3D primitives, and keyframe imagery. This enables, for instance, correctly determining the availability of nearby parking by retrieving likely parking-area memories and re-examining associated keyframes at the relevant timestamps, and aligns with the observed gains on long-horizon textual queries requiring fine-grained visual understanding.
% semantics. For example, when asked “Is there an available parking space nearby?”, ReMEmbR can infer the presence of cars but cannot determine occupancy, while OpenGraph’s coarse object-level memory is similarly insufficient. In contrast, our model retrieves visual memories of likely parking areas and re-examines the associated keyframes at relevant timestamps to infer availability, responding accurately with “Yes, there is an empty spot.” This fine-grained memory-based cross-modal reasoning significantly boosts performance on questions requiring detailed visual understanding over a long horizon.

\begin{figure*}[!t] 
    \vspace{0.2cm}
	\centering
	\includegraphics[width=0.9\linewidth]{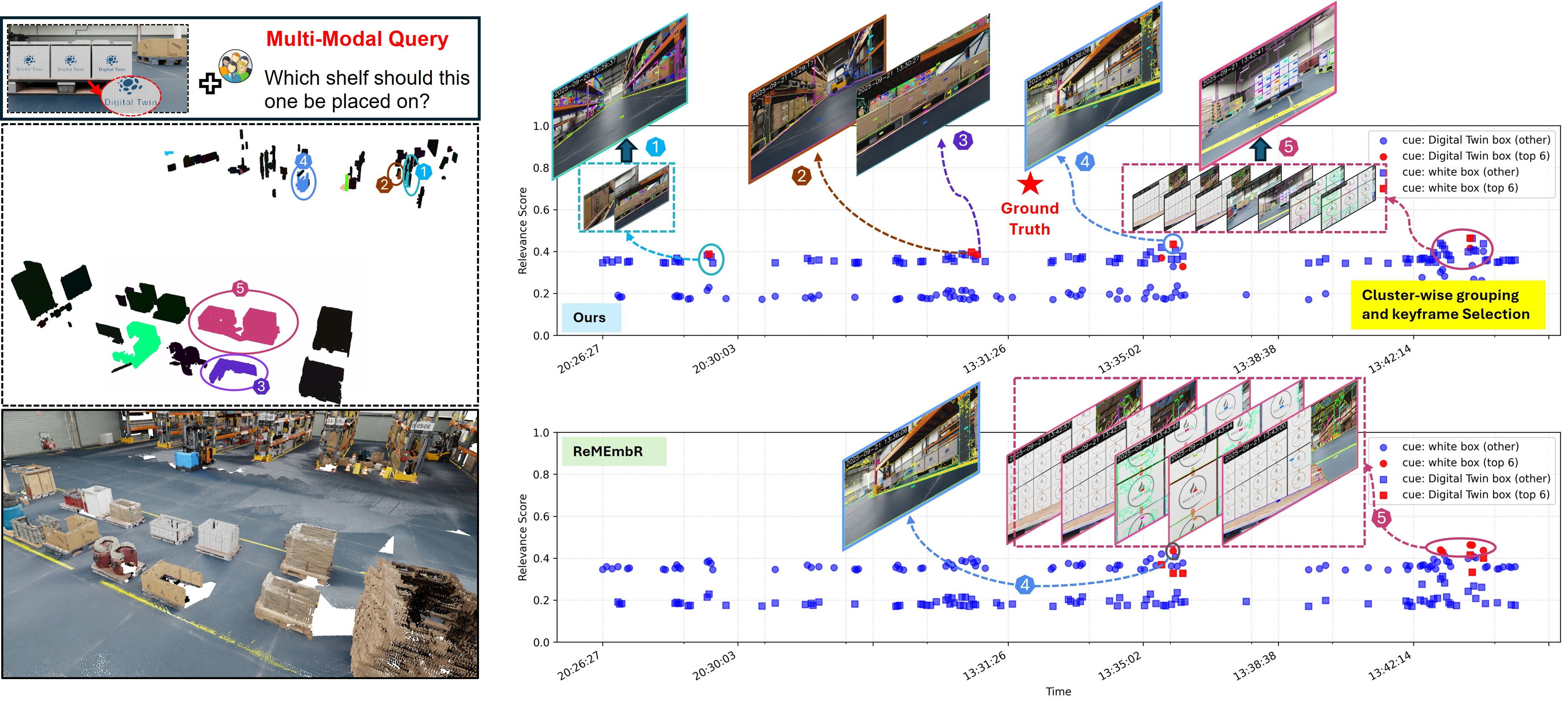}
    \vspace{-0.3cm}
	\caption{\footnotesize \textbf{Qualitative example.} Left: IB-based clustering highlights task-relevant 3D primitives; cues “Digital Twin box” / “white box” guide retrieval, with irrelevant regions masked. Right: our cluster-wise grouping + keyframe selection retrieves diverse, non-redundant memories and correctly grounds Memory 4 (shelf area). ReMEmbR over-samples top-6 cosine hits near 13:43 (Memory 5).}
	\label{fig:warehouse_example}
    \vspace{-0.5cm}
\end{figure*}

\textbf{Qualitative example.} Fig.~\ref{fig:police} illustrates a representative comparison for the query “Where is the nearest police call pole?”. The left panel shows the full reconstructed scene for a 16-minute memory. In the top-left inset, our STaR first reasons over the query and uses “yellow police pole” and “yellow pole” as task cues to retrieves all video-caption memories exceeding a relevance threshold. We then extract the 3D primitives to form a task-specific subset of the scene graph. We then apply IB-based clustering to this subset, masking task-irrelevant primitives in black and highlighting only the most relevant clusters (purple, green, and light-yellow clusters).
By grouping captions via these clusters, our system separates redundant from informative memories. For each cluster, we select the caption with the highest keyword relevance as the representative cue, yielding all memories containing “yellow pole” (right). The green vertical dashed lines mark timestamps where a yellow police pole appears; only the memory around 13:11 explicitly mentions a “yellow pole with a police help sign,” whereas later timestamps (13:15, 13:17, and 13:20) refer to visually similar but semantically incorrect references, such as a “yellow fire hydrant” or a generic “yellow pole,” reflecting perceptual ambiguity.

ReMEmbR, operating purely on textual similarity within a top-6 retrieval window, repeatedly retrieves redundant captions around 13:11 due to high cosine similarity and consequently returns that location as the nearest pole, which is incorrect. In contrast, our approach considers captions and visual evidence across multiple clusters and timestamps: it infers that the “yellow pole” at 13:20, although spatially closer, is not a police call pole, and thus correctly identifies Candidate 2 as the nearest target relative to the robot’s current position (red marker).
% \begin{table}[t!]
%   \centering
%   \setlength{\tabcolsep}{9pt}
%   \renewcommand{\arraystretch}{1.2}
%   \begin{tabular}{@{}llccc@{}}
%     \toprule
%     \textbf{Task} & \textbf{Method} & \textbf{SR}$\uparrow$ & \textbf{Err. (m)}$\downarrow$ & \textbf{Recall}$\uparrow$ \\
%     \midrule
%     \multirow{3}{*}{\textbf{Spatial}}
%       & OpenGraph & 0.27 & 15.5 & 0.30 \\
%       & ReMEmbR   & 0.39 & 12.6 & 0.66 \\
%       & \textbf{STaR}       & \textbf{0.67} & \textbf{6.5} & \textbf{0.73} \\
%     \midrule
%     \multirow{3}{*}{\textbf{Binary}}
%       & OpenGraph & 0.47 & -- & 0.34 \\
%       & ReMEmbR   & 0.44 & -- & 0.36 \\
%       & \textbf{STaR}       & \textbf{0.56} & -- & \textbf{0.57} \\
%     \midrule
%     \multirow{3}{*}{\textbf{Descriptive}}
%       & OpenGraph & 0.0 & -- & 0.27 \\
%       & ReMEmbR   & 0.24 & -- & 0.54 \\
%       & \textbf{STaR}       & \textbf{0.65} & -- & \textbf{0.69} \\
%     \midrule
%     \multirow{3}{*}{\textbf{Mult-modal}}
%       & OpenGraph & 0.19 & -- & 0.30 \\
%       & ReMEmbR   & 0.31 & -- & 0.45 \\
%       & \textbf{STaR}       & \textbf{0.64} & \textbf{--} & \textbf{0.67} \\
%     \bottomrule
%   \end{tabular}
% \caption{\small Comparison on the Warehouse Dataset. We compare our method with two existing approaches: (1) a baseline using OpenGraph’s object captions, and (2) ReMEmbR. We evaluate performance across four task categories.}
%   \label{tab:performance2}
% \end{table}
\begin{table}[t!]
  \centering
  
  \setlength{\tabcolsep}{6pt}
  \renewcommand{\arraystretch}{1.0}
  \begin{tabular}{@{}llccc@{}}
    \toprule
    \textbf{Task} & \textbf{Method} & \textbf{SR}$\uparrow$ & \textbf{Er. (m)}$\downarrow$ & $\mathrm{\textbf{R}}^k$$\uparrow$ \\
    \midrule
    \multirow{3}{*}{\textbf{Spatial}}
      & OpenGraph & 0.27 & 15.5 & 0.30 \\
      & ReMEmbR   & 0.39 & 12.6 & 0.66 \\
      & \textbf{STaR}       & \textbf{0.67} & \textbf{6.5} & \textbf{0.73} \\
    \midrule
    \multirow{3}{*}{\textbf{Text}}
      & OpenGraph & 0.24 & -- & 0.31 \\
      & ReMEmbR   & 0.34 & -- & 0.45 \\
      & \textbf{STaR}       & \textbf{0.63} & -- & \textbf{0.61} \\
    \midrule
    \multirow{3}{*}{\textbf{Mult-modal}}
      & OpenGraph & 0.19 & -- & 0.30 \\
      & ReMEmbR   & 0.31 & -- & 0.45 \\
      & \textbf{STaR}       & \textbf{0.64} & \textbf{--} & \textbf{0.67} \\
    \bottomrule
  \end{tabular}
\caption{\footnotesize Comparison on our WH-VQA dataset. We compare STaR with two baselines: (1) object captions generated by OpenGraph, and (2) ReMEmbR. We evaluate performance across three task categories.}
  \label{tab:performance2}
  \vspace{-0.6cm}
\end{table}

\subsubsection{\textbf{WH-VQA dataset}} WH-VQA is deliberately more challenging than NaVQA: questions demand stronger contextual reasoning and adaptation to variable semantic granularity and overall SR drops for all methods. Even in this setting, our approach performs best (Table~\ref{tab:performance2}), roughly doubling the success rate over the object-caption baseline, clearly outperforming ReMEmbR, and reducing spatial error by about 50–60\%. We also achieve the highest Recall across the Binary, Text, and Multi-modal categories, indicating more accurate and comprehensive memory retrieval.

\noindent\textbf{Qualitative example.} A representative example from WH-VQA (Fig.~\ref{fig:warehouse_example}) involves a multimodal query: “Which shelf should this one be placed on?” paired with the robot’s current observation of the target box. The robot uses cross-modal reasoning to resolve the expression “this one” as a white box labeled “Digital Twin”, and selects “Digital Twin box” and “white box” as textual task cues for memory retrieval. ReMEmbR, restricted to cosine-similarity caption retrieval within a top-6 input window, over-samples redundant memories around 13:43 associated with the white box in the staging area (cluster \#5), diluting information relevant to the shelves. In contrast, our STaR effectively retrieves both shelf-area and staging-area memories, clusters them, and correctly identifies cluster \#4 as the shelf designated for placing the target object.

Although ReMEmbR successfully retrieves relevant memories, it ultimately misidentifies the staging area (cluster \#5) as the final location. When the question difficulty increases---for instance, “How many empty slots remain on that shelf?”---text-only memories are insufficient, since it is infeasible to encode such fine-grained spatial detail in captions alone. STaR responds correctly that one pallet space remains by retrieving the keyframe memory from cluster \#4 and visually inspecting the shelf layout to complete the task. %Please see the project video for more testing results.\footnote{\url{https://drive.google.com/file/d/190kDYuJw-cAHNeeUElmsXRe2yjipcSqi/view?usp=drive_link}}

\subsubsection{\textbf{Implementation Details and On-Device Deployment}}
\begin{figure}[tbp] 
	\centering
\includegraphics[width=1\linewidth]{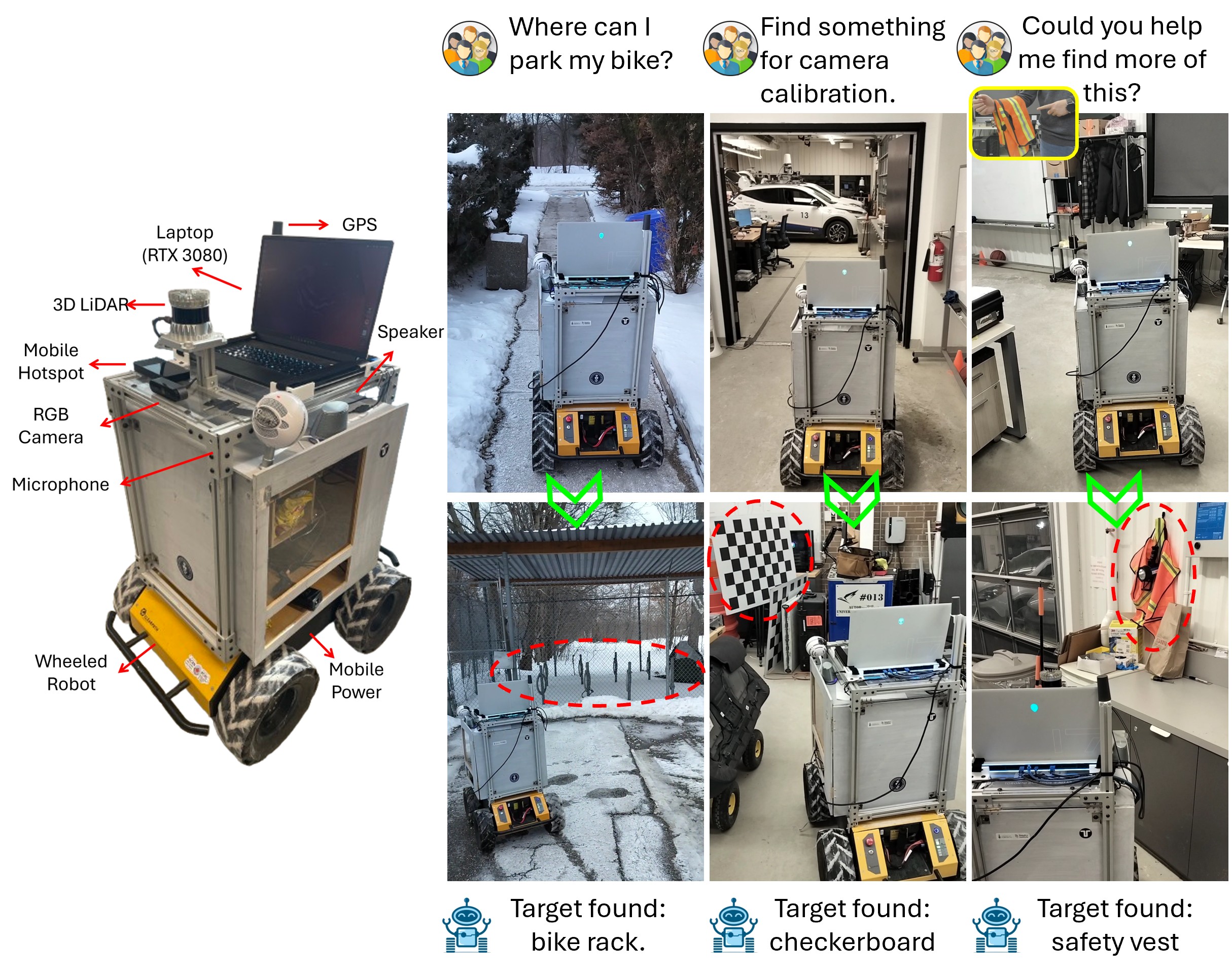}
	\caption{\footnotesize STaR deployed on a Husky robot for indoor and outdoor experiments, supporting both text-based and multimodal queries.}
	\label{fig:hardware}
    \vspace{-0.3cm}
\end{figure}

STaR is deployed in both simulation and real-world settings using a Clearpath Husky robot. Simulation experiments are conducted in Isaac Sim with an RGB-D camera and an Ouster OS1-128 3D LiDAR, while the real robot operates in indoor and outdoor environments using a Logitech RGB camera and the same LiDAR, with poses estimated via Cartographer SLAM. All sensor streams are synchronized and sampled at 10 Hz.
As STaR assumes a pre-explored environment and is not intended for immediate deployment in fully unexplored settings, memory construction is performed via teleoperation with deliberate revisits. During the subsequent query stage, STaR supports both text-only and multimodal queries and is evaluated on 30 tasks of increasing difficulty to assess robustness and end-to-end feasibility under realistic conditions. When a target location is successfully identified, navigation goals are executed using the ROS 2 Nav2 stack on a pre-built map. Representative task examples are shown in Fig.~\ref{fig:hardware}.
Dataset evaluation and simulation experiments run on a server with an RTX 4090, while real-robot memory construction runs on an Alienware X17 R2 laptop with an RTX 3080. We employ a quantized NVILA-Lite-2B model to aggregate video captions (one every 3 s), consuming approximately 7.2 GB of GPU memory. The open-vocabulary perception stack requires an additional 6.2 GB, enabling online multimodal memory construction at 1 Hz on device. 
Our method supports both RGB+LiDAR and RGB-D perception modes. With the longest memory duration (35.9 minutes), multimodal memory footprint is 15.7 MB for video captions, 158.1 MB for 3D primitives, and 4.3 GB for keyframe visual memory, where the latter is dominated by image size ($\sim$1.8 MB/image in NaVQA vs. $\sim$0.9 MB in simulation and real-robot runs, halving storage); keyframe visual memory is stored on local disk and retrieved on demand by the agentic planner via selected timestamps.
\section{CONCLUSION}
This work presents a systematic study of the key challenges faced by robots deployed for extended missions in open environments, including long-horizon memory construction, efficient retrieval, contextual reasoning, and flexible human–robot interaction.
To address these challenges, we proposed STaR, a scalable framework that constructs a task-agnostic multimodal memory capable of handling open-ended user queries.
STaR leverages adaptive retrieval under the IB principle to isolate task-relevant, information-rich, non-redundant memories for contextual reasoning, enabling accurate spatial, temporal, and descriptive outputs.  Experiments on two long-horizon VQA benchmarks demonstrate that STaR significantly outperforms existing approaches. In future work, we plan to extend STaR toward high-level task execution, enabling robots to operate within an agentic workflow that decomposes human instructions into sub-tasks and grounds them precisely in long-term memory.
\bibliographystyle{IEEEtran}%IEEEtran} #ieeetr
\bibliography{root}
\end{document}